\documentclass[journal]{IEEEtran}

\usepackage{cite}

\ifCLASSINFOpdf
   \usepackage[pdftex]{graphicx}
\else
\fi

\usepackage[colorlinks=true,linkcolor=black,citecolor=black]{hyperref}
\usepackage{mathrsfs}
\usepackage{amssymb}
\usepackage{amsmath}
\usepackage{array}
\usepackage{multirow}

\usepackage{color}
\newtheorem{definition}{Definition}

\begin{document}
	
\bstctlcite{BSTcontrol}	
\title{A Fractional-Order Normalized Bouc-Wen Model for Piezoelectric Hysteresis Nonlinearity}

\author{Shengzheng~Kang, Hongtao~Wu, Yao~Li, Xiaolong~Yang, and~Jiafeng~Yao
\thanks{This work was supported in part by the National Key Research and Development Program of China under Grant 2018YFC0309100 and in part by the National Natural Science Foundation of China under Grant 51975277. 
\textit{(Corresponding author: Shengzheng Kang, Xiaolong Yang.)}
	}
\thanks{S. Kang, H. Wu and J. Yao are with the College of Mechanical and Electrical Engineering, Nanjing University of Aeronautics and Astronautics, 29 Yudao Street, Qinhuai District, Nanjing 210016, China (e-mail: kangsz@nuaa.edu.cn; mehtwu@126.com; jiaf.yao@nuaa.edu.cn).}
\thanks{Y. Li is with the Industrial Center, School of Innovation and Entrepreneurship, Nanjing Institute of Technology, 1 Hongjing Avenue, Jiangning District, Nanjing 211167, China (e-mail: liyaokkx@njit.edu.cn).}
\thanks{X. Yang is with the School of Mechanical Engineering, Nanjing University of Science and Technology, 200 Xiaolingwei Street, Xuanwu District, Nanjing 210094, China (e-mail: xiaolongyang@njust.edu.cn).}}

\maketitle

\begin{abstract}
This paper presents a new fractional-order normalized Bouc-Wen (BW) (FONBW) model to describe the asymmetric and rate-dependent hysteresis nonlinearity of piezoelectric actuators (PEAs). In view of the fact that the classical BW (CBW) model is only efficient for the symmetric and rate-independent hysteresis description, the FONBW model is devoted to characterizing the asymmetric and rate-dependent behaviors of the hysteresis in PEAs by adopting an $N$th-order polynomial input function and two fractional operators, respectively. Different from the traditional modified BW models, the proposed FONBW model also eliminates the redundancy of parameters in the CBW model via the normalization processing. By this way, the developed FONBW model has a relatively simple mathematic expression with fewer parameters to simultaneously characterize the asymmetric and rate-dependent hysteresis behaviors of PEAs. Model parameters are identified by the self-adaptive differential evolution algorithm. To validate the effectiveness of the proposed model, a series of model verification and inverse-multiplicative-structure-based feedforward control experiments are carried out on a PEA system. Results show that the proposed model is superior to the CBW model and traditional modified BW model in modeling accuracy and hysteresis compensation.
\end{abstract}


\section{Introduction}
\label{sec_1}
\IEEEPARstart{P}{iezoelectric} actuators (PEAs) are widely used in the applications of precision positioning \cite{gu2016modeling} and microvibration isolation \cite{kang2020fractional,yang2019dynamic}, thanks to their excellent characteristics including high resolution, fast response, and light weight. Unfortunately, the inherent hysteresis nonlinearity of the PEAs often significantly limits their operation accuracy.

To alleviate the hysteresis effects of the PEAs, many compensation methods have been developed \cite{qin2012novel, li2009adaptive,xu2012identification,fan2019design,li2018inverse}. Among the existing methods, feedforward control is commonly used because of its low-cost and no external sensors required \cite{leang2009feedforward}. Generally speaking, its principle consists in establishing the hysteresis model as accurately as possible and then employing the inverse of the model as a compensator to cascade with the controlled system. Motivated by this, there are numerous works in the literature concentrating on hysteresis modeling, such as the Preisach model \cite{tang2015feedforward,song2005tracking}, the Prandtl-Ishlinskii model \cite{gu2013modeling,al2016further}, and the Bouc-Wen (BW) model \cite{rakotondrabe2010bouc}. Compared with the first two models, the BW model has the advantage of computational simplicity, as it only requires one differential equation with a few parameters to describe the hysteresis behavior \cite{ismail2009hysteresis}. However, the classical BW (CBW) model is only efficient for the symmetric and rate-independent hysteresis description \cite{li2016modeling,zhu2016hysteresis}. Consequently, severe modeling errors would occur when the CBW model is utilized to represent the piezoelectric hysteresis nonlinearity which exhibits asymmetric and rate-dependent characteristics. In this regard, various efforts have been recently performed to modify the CBW model. For example, Wang \textit{et al.} \cite{wang2015modeling} and Zhu \textit{et al.} \cite{zhu2012non} introduced a non-odd input function and an asymmetric formula into the CBW model to achieve the asymmetric hysteresis, respectively. Habineza \textit{et al.} \cite{habineza2014multivariable} extended the monovariable generalized BW asymmetric hysteresis model into a multivariable model, where a shape control function is applied to control hysteresis with six shape phases instead of four as in CBW model. For further describing the rate-dependent behavior, Li \textit{et al.} \cite{li2016modeling} developed a Hammerstein structure cascading the asymmetric BW model with the linear dynamics, and Zhu \textit{et al.} \cite{zhu2016hysteresis} proposed a generalized BW model with a frequency factor, which both make it possible to simultaneously characterize the asymmetric and rate-dependent hysteresis effect.

Although exciting results have been obtained in \cite{li2016modeling,zhu2016hysteresis}, they are still not completely satisfactory. For instance, both modified BW hysteresis models in \cite{li2016modeling,zhu2016hysteresis} do not take into account the problem that there exists the inherent redundancy of parameters in the CBW model \cite{ikhouane2007variation}, which makes it more difficult to identify the model parameters. At the same time, redundant parameters also easily lead to the divergence of the numerical solution of the differential equation in the identification process. Moreover, most of the existing modified BW models are limited to an integer-order differential equation. It is known that fractional calculus extends the order of classical calculus from integer domain to complex domain, whose unique nonlocal memory effect provides an excellent potential for the application of hysteresis modeling \cite{zhu2012novel, liu2013hysteresis,zhu2018external,ding2019fractional}. Specifically for the BW model, the fractional calculus could be a good choice for solving the rate-dependent hysteresis problem. However, to the best of authors' knowledge, there have been few works to apply the fractional calculus to the BW model for the hysteresis modeling of PEAs.

In this paper, a novel fractional-order normalized BW (FONBW) model is proposed and investigated to improve the modeling accuracy of the hysteresis nonlinearity in a PEA system. The main contributions include: 1) the redundancy of parameters in the CBW model is considered and eliminated via the normalization processing, such that the model has fewer parameters; 2) an $N$th-order polynomial input function is proposed to describe the hysteresis nonlinearity with the asymmetric behavior; 3) two fractional operators are introduced into the BW model for the first time to characterize the rate-dependent hysteresis effect. The hysteresis parameters of the proposed FONBW model are identified by the self-adaptive differential evolution algorithm on a PEA system. Based on the inverse multiplicative structure of the proposed FONBW model, the feedforward control is also implemented to compensate the hysteresis. Comparative experiments show that the developed model is effective and matches better the system responses than the CBW model and traditional modified BW model \cite{zhu2016hysteresis}.

The remainder of this paper is organized as follows. Section \ref{sec_2} presents the hysteresis modeling of the PEA system, and then the characteristics of the proposed FONBW model are discussed in Section \ref{sec_3}. Afterwards, the model identification, verification, and hysteresis compensation experiments are conducted to verify the effectiveness of the proposed model in Section \ref{sec_4}. Finally, Section \ref{sec_5} concludes this paper.

\section{Hysteresis Modeling of PEA system}
\label{sec_2}

\subsection{Review of the CBW Model}
The CBW model was firstly proposed by Bouc and further modified by Wen for modeling the hysteresis in vibrational mechanics \cite{ismail2009hysteresis}. Owing to the capability of describing many categories of hysteresis and the benefit of simplicity in computing, the CBW model has been extensively applied in piezoelectric hysteresis modeling. In this model, the relationship between the output displacement and input voltage of the PEA system can be expressed in the form of
\begin{eqnarray}
&& H(u,t) = \alpha k u(t) + (1-\alpha) D k h(t) \label{eq_01} \\
&& \dot{h} = D^{-1} \left( A \dot{u} - \beta \left| \dot{u} \right| \left| h \right|^{n-1} h - \gamma \dot{u} \left| h \right|^n \right)  \label{eq_02}
\end{eqnarray}
where $u(t)$ is the input voltage to the PEA; $H(u,t)$ represents the hysteresis output displacement, composed of an elastic term $\alpha k u$ and a purely hysteretic term $(1-\alpha) D k h$ with the parameters $D$, $k$, and $\alpha$; $h(t)$ is an auxiliary hysteresis variable which is the solution of the nonlinear first-order differential equation (\ref{eq_02}), and its shape and magnitude are determined by parameters $A$, $\beta$, $\gamma$ and $n (n \geq 1)$. Through proper choice of these parameter values, a wide range of hysteresis loops can be described. A detailed description of the relationship between the parameters $\alpha, k, D, A, \beta, \gamma, n$ and the hysteresis loop can be found in \cite{baber1980stochastic}.

\begin{figure}[!t]
\centering
\includegraphics[width=2.in]{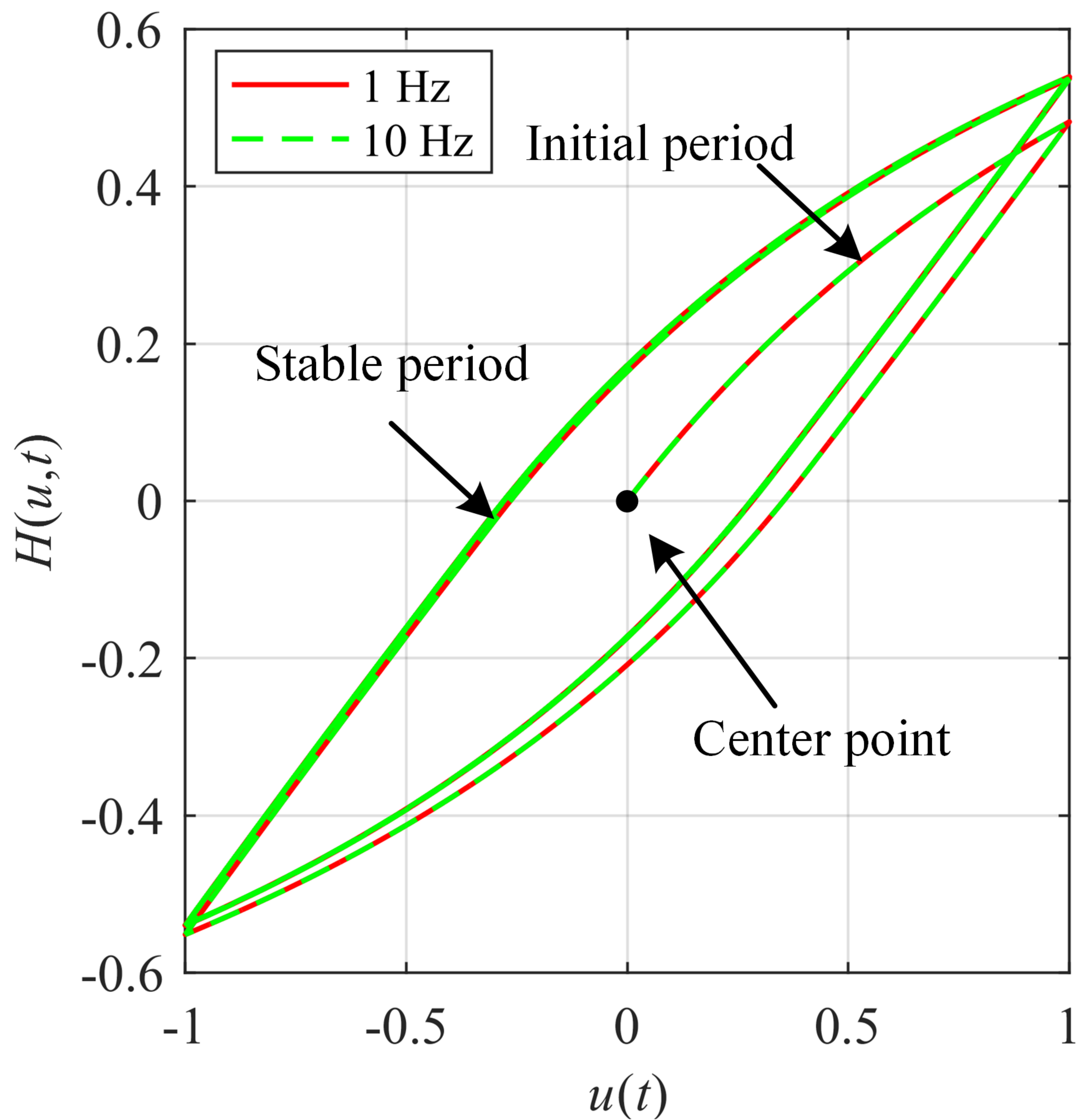}
\caption{Hysteresis curves generated by the CBW model with the different input frequencies.}
\label{fig_01}
\end{figure} 

Fig. \ref{fig_01} shows the hysteresis curves generated by the CBW model with the different input frequencies, if the parameter values are chosen as $\alpha=0.1, k=1, D=1, A=0.7, \beta=0.6, \gamma=0.5$ and $n=1$. It can be found that the hysteresis loop is not only symmetric around its center point, but also does not change with the input frequency. But actually, for the piezoelectric materials, they exhibit the asymmetric and rate-dependent behaviors \cite{hassani2014survey}, which are unavailable for the CBW model. Therefore, to further improve the modeling accuracy, it is necessary to make a modification.

\subsection{Normalization Processing}
Although several works \cite{wang2015modeling,zhu2012non, li2016modeling, zhu2016hysteresis} have been devoted to modifying the CBW model to guarantee the asymmetric or rate-dependent characteristic, they generally still suffer from the redundancy of parameters, which originally exists in the CBW model \cite{ikhouane2007variation}. To facilitate the description of parameter redundancy, without loss of generality, taking two different sets of parameters for example: $n_1=n_2=n$, $A_1=A_2$, $\alpha_1=\alpha_2$, $k_1=k_2$, $\beta_2=c^n \beta_1$, $\gamma_2=c^n \gamma_1$, $D_2=c D_1$, where $c$ is a positive constant, and the initial condition $h_1(0) =h_2(0)=0$. Thus, the CBW model in Eqs. (\ref{eq_01}) and (\ref{eq_02}) can be rewritten in the following two forms:
\begin{equation} \label{eq_03}
\begin{cases}
H_1(u,t)=\alpha_1 k_1 u(t) + (1-\alpha_1) D_1 k_1 h_1(t) \\
\dot{h}_1 = D^{-1}_1 \left( A_1 \dot{u} - \beta_1 \left| \dot{u} \right| \left| h_1 \right|^{n-1} h_1 - \gamma_1 \dot{u} \left| h_1 \right|^n \right)
\end{cases}
\end{equation}
and
\begin{equation} \label{eq_04}
\begin{cases}
H_2(u,t)=\alpha_2 k_2 u(t) + (1-\alpha_2) D_2 k_2 h_2(t) \\
\dot{h}_2 = D^{-1}_2 \left( A_2 \dot{u} - \beta_2 \left| \dot{u} \right| \left| h_2 \right|^{n-1} h_2 - \gamma_2 \dot{u} \left| h_2 \right|^n \right) 
\end{cases}
\end{equation}

According to the given parameters, Eq. (\ref{eq_04}) can be transferred into
\begin{equation} \label{eq_05}
\begin{cases}
H_2(u,t)=\alpha_1 k_1 u(t) + c (1-\alpha_1) D_1 k_1 h_2(t) \\
\dot{h}_2 = c^{-1} D^{-1}_1 \left( A_1 \dot{u} - c^n \beta_1 \left| \dot{u} \right| \left| h_2 \right|^{n-1} h_2 - c^n \gamma_1 \dot{u} \left| h_2 \right|^n \right) 
\end{cases}
\end{equation}
Letting $h_a(t)=c h_2(t)$, Eq. (\ref{eq_05}) is then expressed as
\begin{equation} \label{eq_06}
\begin{cases}
H_2(u,t)=\alpha_1 k_1 u(t) + (1-\alpha_1) D_1 k_1 h_a(t) \\
\dot{h}_a = D^{-1}_1 \left( A_1 \dot{u} - \beta_1 \left| \dot{u} \right| \left| h_a \right|^{n-1} h_a - \gamma_1 \dot{u} \left| h_a \right|^n \right) 
\end{cases}
\end{equation}
It is obviously found from Eqs. (\ref{eq_03}) and (\ref{eq_06}) that both models exactly deliver the same hysteresis loop $H(u,t)$ for any input signal $u(t)$. That means the input-output behavior of the CBW model is not determined by a unique set of parameters $\left\lbrace \alpha, k, D, A, \beta, \gamma, n \right\rbrace$. For this reason, it is necessary to normalize the CBW model, making the parameters defined in a unique way.

To this end, define the constants $h_0=\sqrt[n]{\frac{A}{\beta+\gamma}}$, $\rho=\frac{A}{D h_0}$, $\sigma=\frac{\beta}{\beta+\gamma}$, $k_u=\alpha k$, $k_h=(1-\alpha) D k h_0$ and the parameter variable $\hbar(t)= \frac{h(t)}{h_0}$. Substituting them into Eqs. (\ref{eq_01}) and (\ref{eq_02}) yields
\begin{eqnarray}
&& H(u,t) = k_u u(t) + k_h \hbar(t) \label{eq_07} \\
&& \dot{\hbar} = \rho \left( \dot{u} - \sigma \left| \dot{u} \right| \left| \hbar \right|^{n-1} \hbar + (\sigma - 1) \dot{u} \left| \hbar \right|^n \right)  \label{eq_08}
\end{eqnarray}
which are the so-called normalized BW (NBW) model \cite{ikhouane2007variation}. Note that the normalized form of the CBW model is exactly equivalent to its standard form (\ref{eq_01}) and (\ref{eq_02}), if the initial condition $\hbar(0)=\frac{h(0)}{h_0}$. Moreover, it also has the advantage of having only five parameters instead of the seven parameters for the standard form, which make it easier to be identified. Based on the normalized form (\ref{eq_07}) and (\ref{eq_08}), next, a modified FONBW model will be developed to describe the hysteresis effect in PEAs with asymmetric and rate-dependent characteristics.

\begin{figure}[!t]
\centering
\includegraphics[width=2.in]{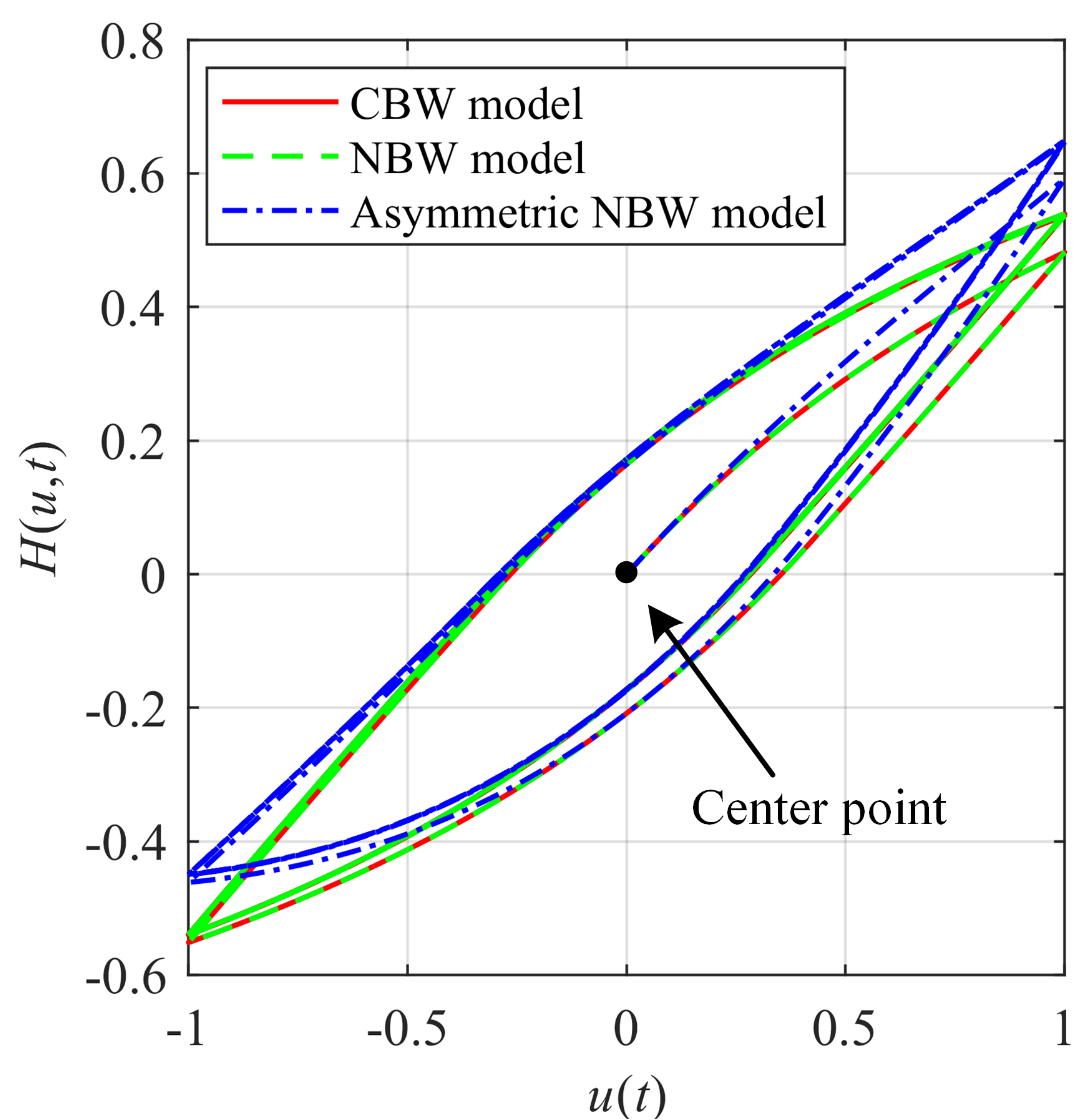}
\caption{Comparison of the hysteresis curves generated by the different BW models.}
\label{fig_02}
\end{figure} 

\subsection{Proposed FONBW Model}
Inspired by the work in \cite{gu2013modeling,bashash2008polynomial}, to describe the hysteresis in PEAs which exhibits asymmetric characteristic, an $N$th-order polynomial input function with the memoryless and locally Lipschitz continuous properties is utilized by modifying Eq. (\ref{eq_07}) as follows:
\begin{equation} \label{eq_09}
H(u,t) = g(u,t) + k_h \hbar(t)  
\end{equation}
with
\begin{equation} \label{eq_010}
g(u,t) = k_{u1} u(t) + k_{u2} u^2(t) + \cdots + k_{uN} u^N(t) 
\end{equation}
where $k_{ui} (i=1,2,\cdots,N)$ is the coefficient of the polynomial function $g(u,t)$. Combining Eqs. (\ref{eq_09}) and (\ref{eq_010}) with (\ref{eq_08}), the asymmetric NBW model is obtained. To validate the effectiveness of the normalization and asymmetric processing, the output of the asymmetric NBW model under the sinusoidal input signal is depicted in Fig. \ref{fig_02} by simulation. The parameters of the asymmetric NBW model are chosen as $N=3, k_{u1}=0.1, k_{u2}=0.1, k_{u3}=0.01$, and the other parameters are consistent with those of the CBW model in Fig. \ref{fig_01}. As a comparison, the outputs of the CBW and NBW models are also shown in Fig. \ref{fig_02}. It can be found that, compared with the CBW model, the NBW model has the same hysteresis loop with less parameters, and the asymmetric NBW model enables to describe the asymmetric behavior.

Furthermore, since the fractional calculus \cite{zhu2012novel,liu2013hysteresis,zhu2018external,ding2019fractional} is generally recognized as an effective choice to describe the hysteresis loops with rate-dependent characteristic, the idea is also introduced here to develop the FONBW model. To this end, the nonlinear hysteresis effect (\ref{eq_08}) is extended from the integer order to the fractional order as
\begin{equation} \label{eq_011}
\mathscr{D}^{\lambda_2} \hbar = \rho \left( \mathscr{D}^{\lambda_1} u - \sigma \left| \mathscr{D}^{\lambda_1} u \right| \left| \hbar \right|^{n-1} \hbar + (\sigma - 1) \mathscr{D}^{\lambda_1} u \left| \hbar \right|^n \right)  
\end{equation}
where the operator $\mathscr{D}^{\lambda_i}(i=1,2)$ denotes the fractional derivative with the order $0 < \lambda_i \leq 1$. Here, if taking into account the Gr\"{u}nwald-Letnikov's definition (See Definition \ref{def_1} in Appendix \ref{app_1}.) for the fractional-order derivatives, one can know that the calculation of the hysteresis variable $\hbar(t)$ in Eq. (\ref{eq_011}) not only depends on the current input voltage, but also requires all of its history states due to the existence of summation operation (See Appendix \ref{app_1} for the proof.). That is to say, the nonlocal memory effect of the fractional calculus is essentially beneficial for the description of the inherent rate-dependent property of the piezoelectric hysteresis. To the best of authors' knowledge, this fact has not been recognized in any other modified BW models. Thus, the proposed FONBW model is finally synthesized by Eqs. (\ref{eq_09})-(\ref{eq_011}).

It is worth mentioning that, since the proposed FONBW model is developed on the CBW model, the purpose of normalization processing is to make the hysteresis shapes of the FONBW model determined by a unique parameter set, reducing the identification complexity and avoiding the parameters redundancy. Although the number of parameters in the proposed FONBW model is ultimately bigger than that in the CBW model, it is beneficial for the description of asymmetric and rate-dependent hysteresis effect. Actually, compared with the traditional modified BW models \cite{li2016modeling,zhu2016hysteresis}, the proposed model indeed has an advantage in terms of the number of parameters, which will be seen in Table \ref{tab_02}.

\begin{figure}[!t]
	\centering
	\includegraphics[width=3.5in]{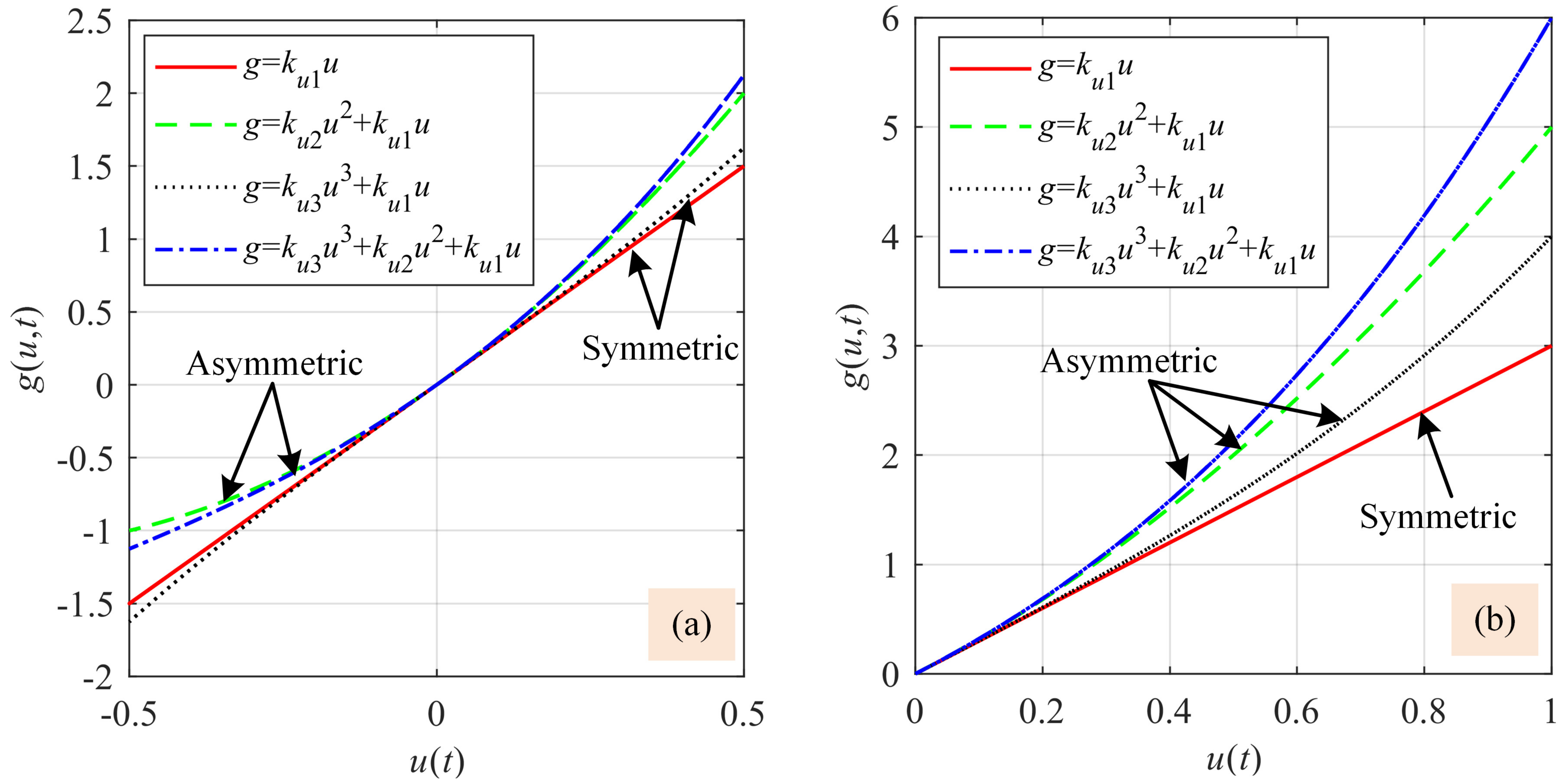}
	\caption{Comparison of different polynomial functions with symmetric input voltage about the origin (a) and positive input voltage (b). It can be seen from (a) that, $g = k_{u1} u + k_{u2} u^2 + k_{u3} u^3$ and $g = k_{u1} u + k_{u2} u^2$ are asymmetric about the center point of $u$, while $g = k_{u1} u + k_{u3} u^3$ and $g = k_{u1} u$ are symmetric. Similarly, from (b), one can see that $g = k_{u1} u + k_{u2} u^2 + k_{u3} u^3$, $g = k_{u1} u + k_{u3} u^3$ and $g = k_{u1} u + k_{u2} u^2$ are asymmetric about the center point of $u$, while $g = k_{u1} u$ is symmetric.}
	\label{fig_03}
\end{figure} 

\section{Characteristics of Proposed FONBW Model}
\label{sec_3}
As stated in the previous section, the proposed FONBW model not only inherits the advantages of the CBW model, but also makes it possible to simultaneously achieve the asymmetric and rate-dependent characteristics with the relative simple mathematic expression. In this section, these notable characteristics will be further discussed in combination with the model parameters.

\subsection{Asymmetric Characteristic}

It has been disclosed that the asymmetric characteristic of the proposed FONBW model results from the selection of polynomial function $g(u,t)$. In order to intuitively describe the influence of the order of $g(u,t)$ on the asymmetric characteristic, without loss of generality, taking the order $N=3$ and the coefficients $k_{u1}=3, k_{u2}=2, k_{u3}=1$ in Eq. (\ref{eq_010}) for example, the corresponding curve is demonstrated in Fig. \ref{fig_03}, where the other curves are shown for the purpose of comparison.

As can be seen, when the input voltage is symmetrical about the origin (shown in Fig. \ref{fig_03}(a)), non-odd term needs to exist in $g(u,t)$ to produce the asymmetric behavior, while for the positive input voltage (shown in Fig. \ref{fig_03}(b)), any second- or higher-order polynomial function can result in the asymmetric characteristic. Moreover, the higher order of $g(u,t)$, the more shapes of hysteresis description, but the more identification complexity. Therefore, the order of $g(u,t)$ can be chosen by weighing the shapes of the hysteresis and number of parameters.

It should be noted that, the polynomial functions $g(u,t) = k_{u1} u(t) + k_{u3} u^3(t)$ and $g(u,t) = k_{u1} u(t) + k_{u2} u^2(t)$ shown in Fig. \ref{fig_03} are exactly the types adopted in the modified BW models \cite{li2016modeling, wang2015modeling} to characterize the asymmetric hysteresis, respectively. Nevertheless, the polynomial input function in this work makes the proposed FONBW model accommodate to describe more hysteresis shapes with the asymmetric behavior.

\subsection{Rate-Dependent Characteristic}

\begin{figure}[!t]
\centering
\includegraphics[width=3.5in]{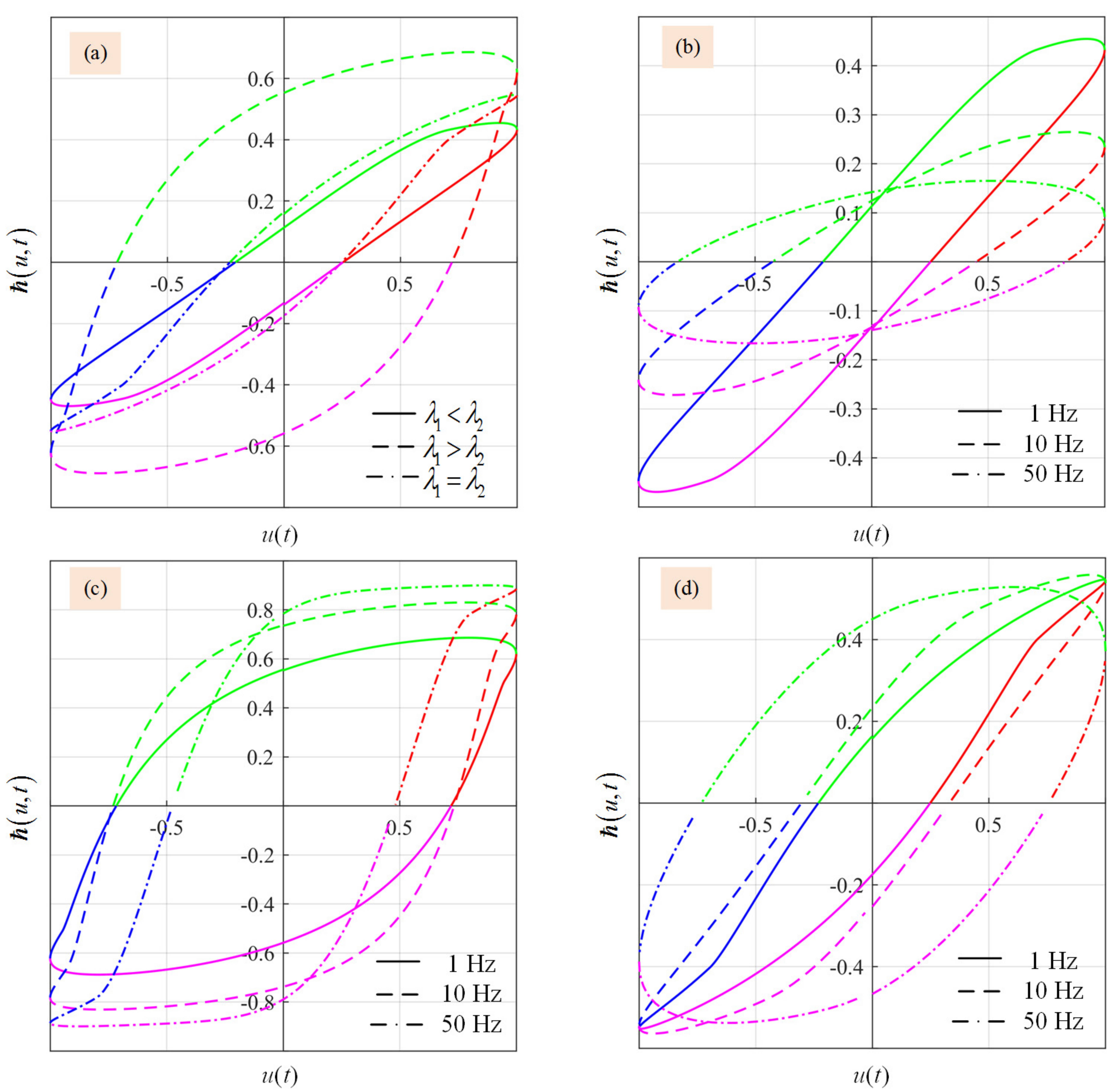}
\caption{Comparison of the hysteresis loops under the different simulation conditions: (a) for 1-Hz input frequency; (b) for $\lambda_1< \lambda_2$; (c) for $\lambda_1>\lambda_2$; (d) for $\lambda_1= \lambda_2$. Moreover, the red color represents the case of $\hbar \geq 0, \mathscr{D}^{\lambda_1} u \geq 0$, the green color for $\hbar \geq 0, \mathscr{D}^{\lambda_1} u \leq 0$, the blue color for $\hbar \leq 0, \mathscr{D}^{\lambda_1} u \leq 0$ and the magenta color for $\hbar \leq 0, \mathscr{D}^{\lambda_1} u \geq 0$.}
\label{fig_04}
\end{figure} 

Thanks to the unique nonlocal memory effect of fractional calculus, the proposed FONBW model enables to describe the rate-dependent hysteresis in PEAs. Consequently, to more clearly seen the nature of the hysteresis model, Eq. (\ref{eq_011}) is broken into following four fractional differential equations:
\begin{eqnarray}
\frac{\mathscr{D}^{\lambda_2} \hbar}{\mathscr{D}^{\lambda_1} u} = \rho (1-\hbar^n)  \qquad \hbar \geq 0, \mathscr{D}^{\lambda_1} u \geq 0 \label{eq_012} \\
\frac{\mathscr{D}^{\lambda_2} \hbar}{\mathscr{D}^{\lambda_1} u}  = \rho \left( 1+(2 \sigma -1)\hbar^n \right) \qquad \hbar \geq 0, \mathscr{D}^{\lambda_1} u \leq 0 \label{eq_013} \\
\frac{\mathscr{D}^{\lambda_2} \hbar}{\mathscr{D}^{\lambda_1} u}  = \rho \left( 1-(-1)^n \hbar^n \right) \qquad \hbar \leq 0, \mathscr{D}^{\lambda_1} u \leq 0 \label{eq_014}  \\
\frac{\mathscr{D}^{\lambda_2} \hbar}{\mathscr{D}^{\lambda_1} u}  = \rho \left( 1+(-1)^n (2 \sigma -1)\hbar^n \right) \quad \hbar \leq 0, \mathscr{D}^{\lambda_1} u \geq 0 \label{eq_015} 
\end{eqnarray}
It can be found that each equation represents a part of the hysteresis loop, i.e., Eq. (\ref{eq_012}) for the positive ascending branch, Eq. (\ref{eq_013}) for the positive descending branch, Eq. (\ref{eq_014}) for the negative descending branch and Eq. (\ref{eq_015}) for the negative ascending branch. They can be easily solved with the Simulink software by choosing proper parameter values, and the solutions are depicted in Fig. \ref{fig_04} with different simulation conditions. 

Note that the parameters $\rho,\sigma$ and $n$ also affect the hysteresis shape of the FONBW model, but they are not the cause resulting in the rate-dependent characteristic. The influences of those three parameters have been discussed from a mathematical point of view in \cite{ikhouane2007variation}. Thus, only the fractional orders $\lambda_1$ and $\lambda_2$ are considered here. In the whole simulations for Fig. \ref{fig_04}, the parameters $\rho, \sigma$ and $n$ are all set to be 1, but $\lambda_1, \lambda_2$ are taken into account the following three cases: 1) $\lambda_1=0.5, \lambda_2=0.8$; 2) $\lambda_1=0.8, \lambda_2=0.5$; 3) $\lambda_1=0.5, \lambda_2=0.5$. From Fig. \ref{fig_04}, it is seen that a large number of hysteresis shapes can be described by varying the parameters $\lambda_1$ and $\lambda_2$. Moreover, with the input frequency increased, the hysteresis shape also changes regularly, which validates the rate-dependent characteristic of the proposed FONBW model.

\subsection{Inverse Multiplicative Structure}
\begin{figure}[!t]
\centering
\includegraphics[width=2.in]{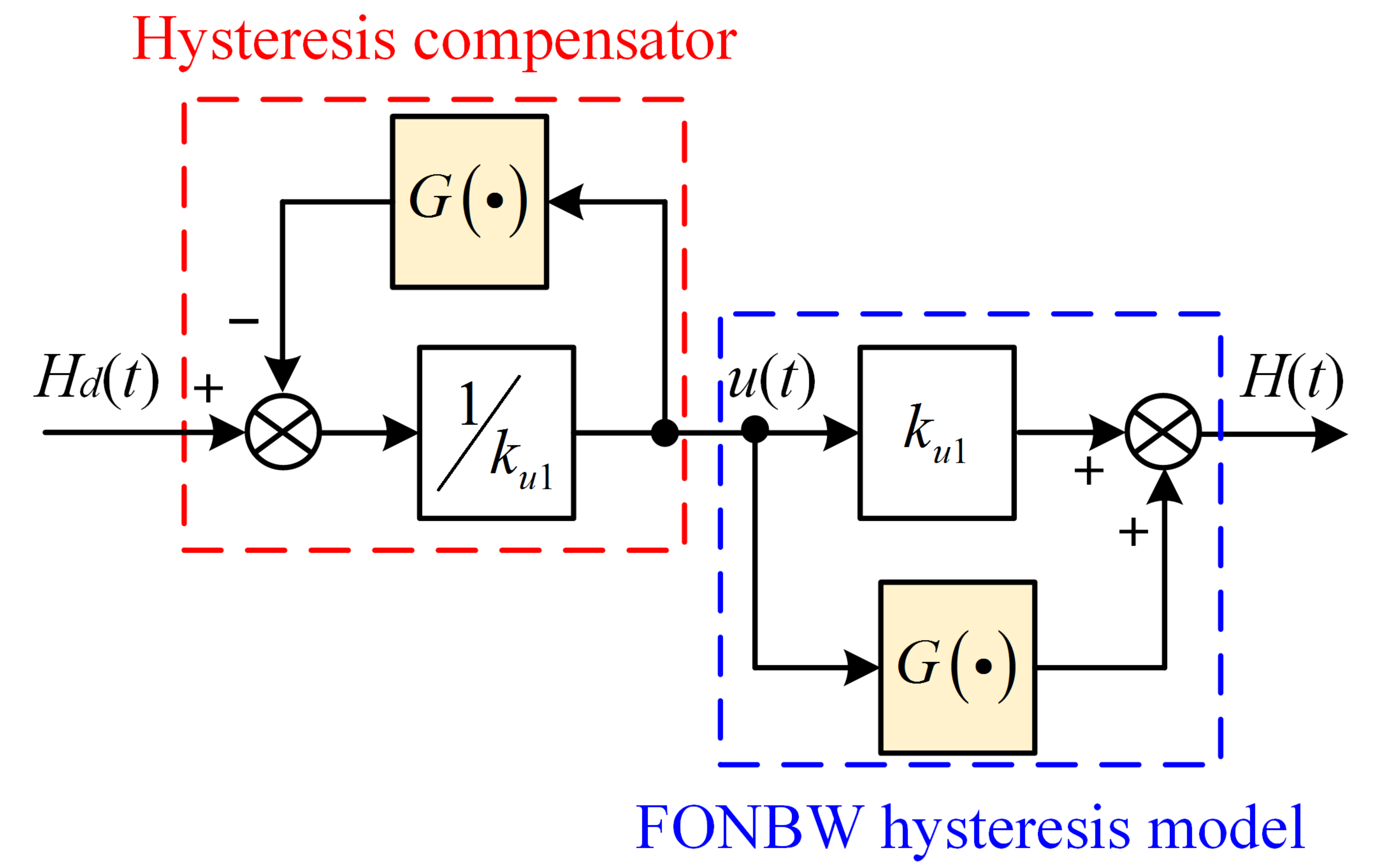}
\caption{Block diagram of the inverse multiplicative structure of the proposed FONBW model.}
\label{fig_05}
\end{figure} 

Besides the aforementioned characteristics, the proposed FONBW model also exhibits an attractive potential in the hysteresis compensation due to its simple inverse multiplicative structure.

Inspired by the idea of utilizing the CBW model to compensate static hysteresis directly without solving the model inversion \cite{rakotondrabe2010bouc}, the inverse multiplicative structure of the proposed FONBW model is obtained from Eqs. (\ref{eq_09})-(\ref{eq_011}) by extracting the input voltage $u(t)$ as
\begin{equation} \label{eq_016}
u(t) = \frac{1}{k_{u1}} \left( H_d(t)-G(u,t) \right) 
\end{equation}
with
\begin{equation} \label{eq_017}
G(u,t) = k_h \hbar(t)+k_{u2} u^2(t)+\cdots + k_{uN} u^N(t)
\end{equation}
where $H_d(t)$ is the given desired displacement. From the control point of view, Eq. (\ref{eq_016}) can be directly employed as a compensator for the hysteresis nonlinearity with the asymmetric and rate-dependent behaviors. The corresponding block diagram is depicted in Fig. \ref{fig_05}. Since $k_{u1}$ and $G(u,t)$ are already known during the modeling and identification, no additional calculation for the inverse model is required, which finally makes the compensator simple to be implemented.

\section{Model Identification and Verification}
\label{sec_4}
In this section, the parameters of the proposed FONBW model are identified on a real PEA system, and its effectiveness is also verified by comparative experiments.

\subsection{Experimental Setup}
\begin{figure}[!t]
\centering
\includegraphics[width=3.5in]{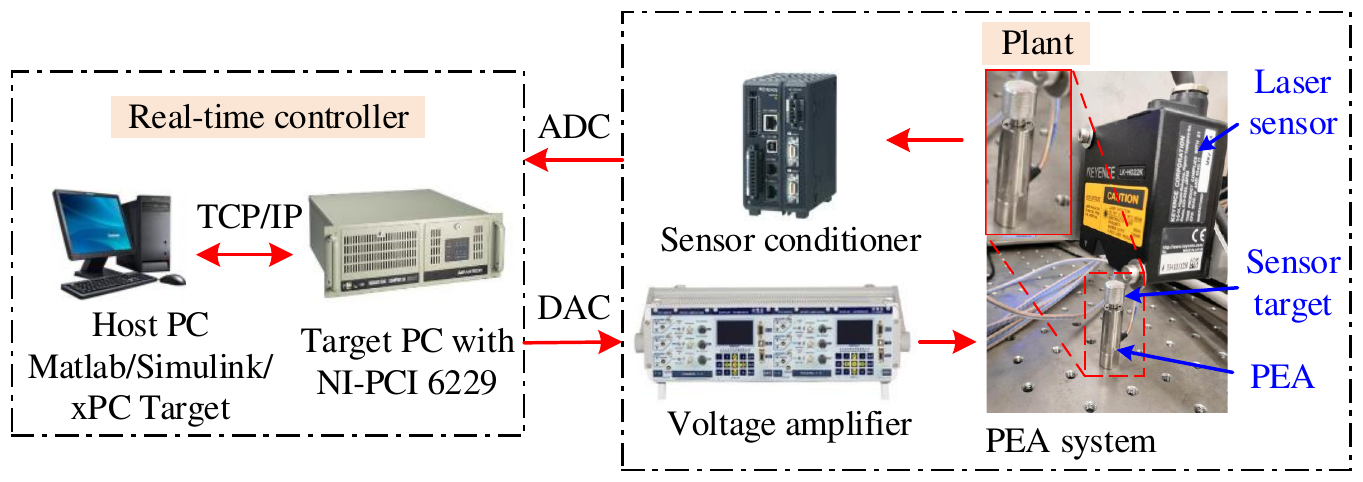}
\caption{Experimental setup for hysteresis modeling.}
\label{fig_06}
\end{figure}

The experimental setup is shown in Fig. \ref{fig_06}. The adopted PEA (model Pst120/7/20VS12 with maximal voltage of 120 V, Core-Tomorrow Co., China) is actuated through a voltage amplifier (model E00.D6 with 6 channels from the Core-Tomorrow Co.). Its output displacement is real-time measured by a laser displacement sensor (model LK-H022, Keyence Co., Japan). The sensor output voltage signal is passed through a signal conditioner (model LK-G5001 from the Keyence Co.), and then acquired by the A/D channel of a data acquisition card (model PCI-6229 with 16-bit A/D and D/A converters from NI Co.). The voltage control signal is produced by the D/A channel and then amplified 12 times via the voltage amplifier to drive the PEA. Programs are developed with Matlab/Simulink software on a host PC and downloaded through the TCP/IP mode to a target PC. In the process of experiments, the sampling frequency is set to be 10 kHz. 

Moreover, the toolbox FOMCOM \cite{tepljakov2011fomcom} is used here to implement the fractional-order terms in the proposed FONBW model, which are approximated by a fifth-order refined Oustaloup filter and the frequency range is set as 0.01-1000 rad/s. Through a tradeoff between the model accuracy and the identification complexity, the order of polynomial function $g(u,t)$ is chosen as $N=3$.

\subsection{Parameter Identification}
\begin{figure}[!t]
	\centering
	\includegraphics[width=2.in]{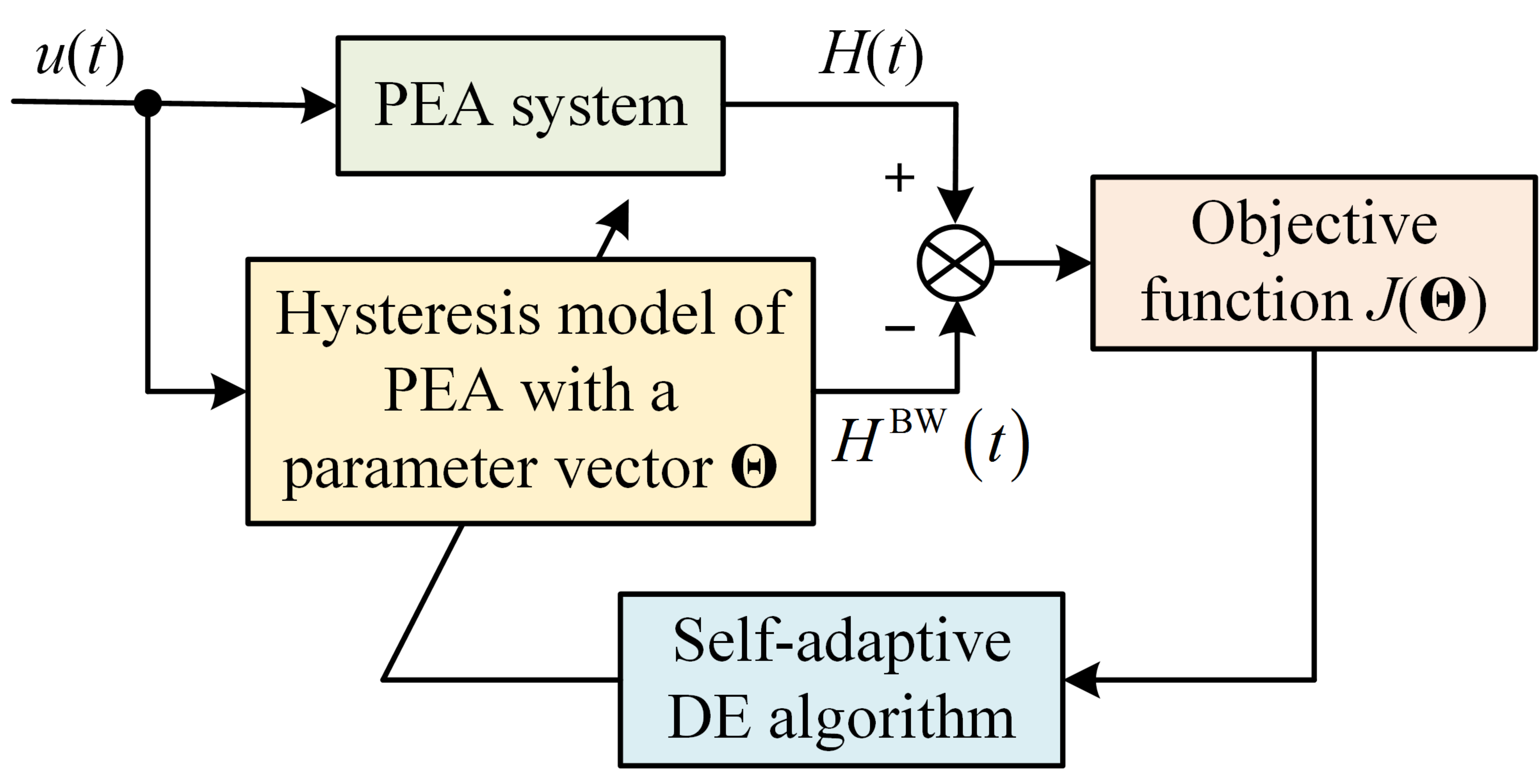}
	\caption{Flow chart of the parameter identification.}
	\label{fig_07}
\end{figure} 

In the traditional modified BW models \cite{li2016modeling, zhu2016hysteresis}, the rate-dependent property is produced by the combination of linear dynamics and rate-independent hysteresis. Due to the strong coupling effect of the hysteresis nonlinearity and dynamics behaviors in the PEA system, it is not easy to effectively identify the parameters of coupled dynamics through conventional two decoupled steps, i.e., the parameters for the linear and hysteresis components are individually identified. Fortunately, considering the operating frequency range ($\leq 20$ Hz) of PEA system in this work, the influence of linear dynamics is not dominant and is thus unnecessary to be identified thanks to the inherent rate-dependent property of the FONBW model. 

Nevertheless, the identification of hysteresis models is still a challenging task, and many algorithms including the least mean square, genetic algorithm, and particle swarm optimization have been developed to solve this problem \cite{hassani2014survey}. In this work, a self-adaptive differential evolution (DE) algorithm \cite{wang2015modeling} featuring simple evolutionary scheme and few tunable parameters is adopted as an illustration. The model identification process is carried out offline and summarized in Fig. \ref{fig_07}, where the root-mean-square (RMS) error is employed as the objective function for reflecting the modeling deviation
\begin{equation} \label{eq_018}
J(\boldsymbol{\Theta}) = \sqrt{\frac{1}{M} \sum_{i=1}^{M} (H_i-H_{i}^{BW})^2}
\end{equation}
where $\boldsymbol{\Theta}=\left[ k_{u1}, k_{u2}, k_{u3}, k_h, \rho, \sigma, n, \lambda_1, \lambda_2 \right]^{\mathrm{T}}$ is the parameter vector to be identified, $M$ is the number of sampling points, $H_i$ is the $i$th measured displacement, and $H_{i}^{BW}$ is the $i$th output displacement generated by the FONBW model. Besides, a variable-amplitude variable-frequency input voltage signal $u(t)$ is also chosen as an illustration for the identification
\begin{equation} \label{eq_019}
u(t) = 60 e^{-0.13 t} \left[ \cos \left( 3 \pi t e^{-0.09 t} -3.15 \right) + 1 \right] \mathrm{V}
\end{equation}

\renewcommand\arraystretch{1.5}
\begin{table}[!t]
	\centering
	\caption{Parameter Identification Results of the PEA System (Unit: SI)}
	\label{tab_01}
	\begin{tabular}{p{0.3cm}<{\centering} p{1.8cm}<{\centering} | p{0.3cm}<{\centering} p{1.8cm}<{\centering} | p{0.3cm}<{\centering} p{1.8cm}<{\centering}}
		\hline
		\hline
		\multicolumn{2}{c|}{FONBW model} & \multicolumn{2}{c|}{CBW model} & \multicolumn{2}{c}{Model in \cite{zhu2016hysteresis}} \\ 
		\hline
		$k_{u1}$ & 0.1811 & $k_a$ & 0.1547 & $m_0$ & 0.1026 \\
		$k_{u2}$ & \tiny $-1.4037 \times 10^{-4}$ \normalsize  & $k_b$ & \tiny $-3.6660 \times 10^{5}$ \normalsize & $c_0$ & \tiny $2.5820 \times 10^{2}$ \normalsize \\
		$k_{u3}$ & \tiny $-7.7154 \times 10^{-8}$ \normalsize & $D$ & 0.5552 & $k_0$ & \tiny $1.5567 \times 10^{5}$ \normalsize\\
		$k_h$ & \tiny $-3.2719 \times 10^{4}$ \normalsize & $A$ & \tiny $6.1987 \times 10^{-7}$ \normalsize & $k_1$ & \tiny $4.3915 \times 10^{-7}$ \normalsize \\
		$\rho$ & \tiny $6.4808 \times 10^{-7}$ \normalsize & $\beta$ & 0.0364 & $x_0$ & 0.0000\\
		$\sigma$ & \tiny $1.3039 \times 10^{5}$ \normalsize & $\gamma$ & 0.0272 & $\tau$ & \tiny $2.0408 \times 10^{-5}$ \normalsize \\
		$n$ & 2.0006 & $n$ & 1.0003 & $A$ & -0.0068\\
		$\lambda_1$ & 0.9557 & - & - & $\beta$ & 0.0457 \\
		$\lambda_2$ & 0.6220 & - & - & $\gamma$ & -0.0255\\
		- & - & - & - & $\delta$ & -0.0024 \\
		- & - & - & - & $n$ & 1.0483 \\
		\hline
		\multicolumn{2}{c|}{RMS error: 0.23 $\mu$m} & \multicolumn{2}{c|}{RMS error: 0.49 $\mu$m} & \multicolumn{2}{c}{RMS error: 0.29 $\mu$m} \\
		\hline
		\hline
	\end{tabular}
\end{table}

\begin{figure}[!t]
\centering
\includegraphics[width=3.5in]{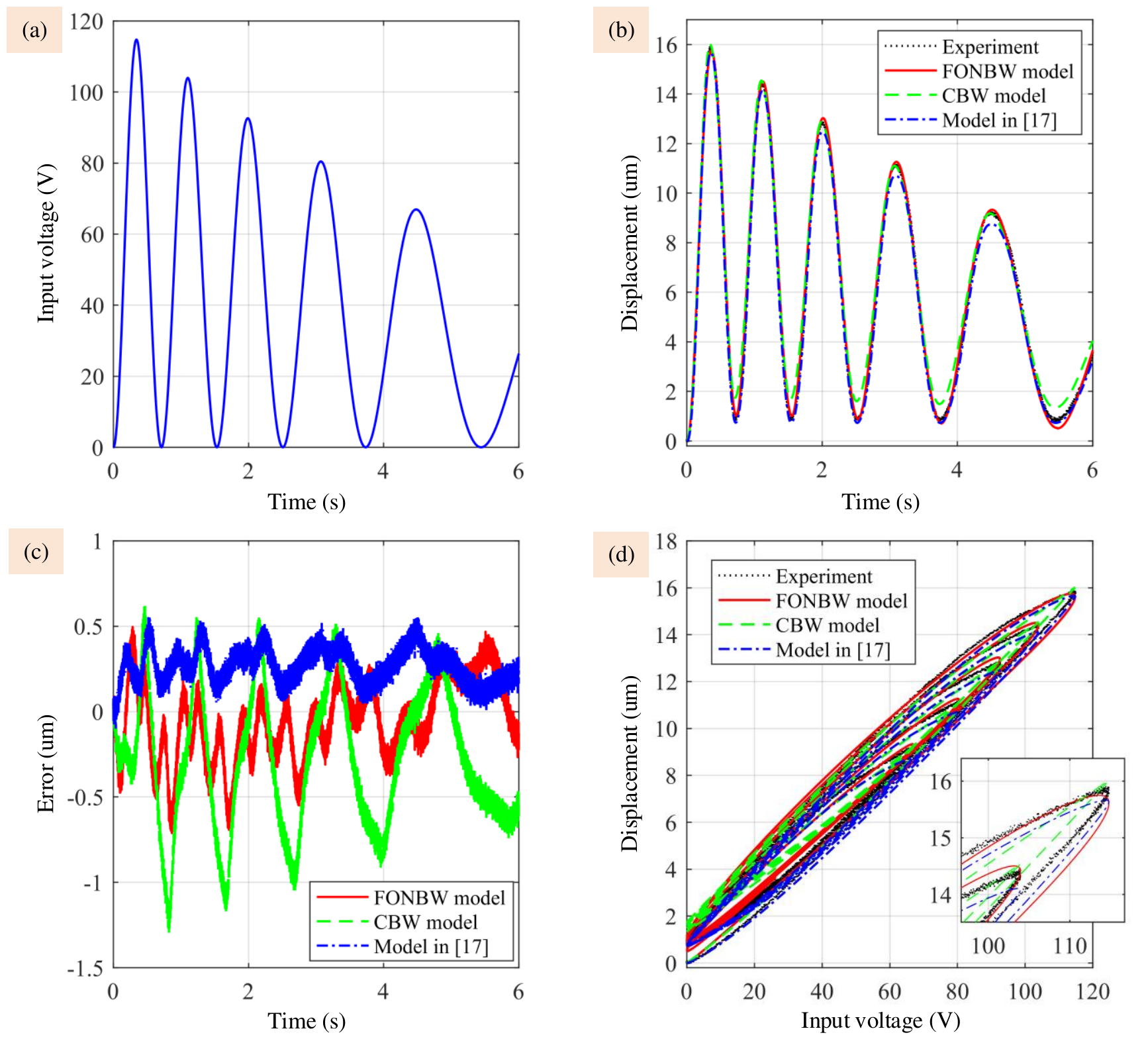}
\caption{Identification results of the proposed FONBW and traditional BW models. (a) Input voltage. (b) Output displacements. (c) Model output errors. (d) Hysteresis loops.}
\label{fig_08}
\end{figure}

The identified model parameters and the RMS error are shown in Table \ref{tab_01}. With the obtained model, Fig. \ref{fig_08} illustrates the model output and experimental results. As a contrast, the parameters of the CBW model and the modified BW model in \cite{zhu2016hysteresis} are also identified in the same way. The model in \cite{zhu2016hysteresis} is introduced here as
\begin{equation} \label{eq_020}
\begin{cases}
m_0 \ddot{x} + c_0 \dot{x} +k_0 (x-x_0)= \frac{k_1}{\tau} e^{-t/\tau}u + h  \\
\dot{h} = A \dot{u} - \beta \left| \dot{u} \right| \left| h \right|^{n-1} h - \gamma \dot{u} \left| h \right|^n + \delta u \mathrm{sgn}(\dot{u}) 
\end{cases}
\end{equation}
where $x$ and $x_0$ are the output displacement and the initial displacement of PEA system, respectively; $m_0$, $c_0$ and $k_0$ are the mass, damping and stiffness of PEA system, respectively; $k_1, \tau, A, \beta, n, \gamma$ and $\delta$ are constants determining the shape of hysteresis loop. As reported, this modified BW model can describe the piezoelectric hysteresis with asymmetric and rate-dependent properties, which will make the comparisons convincing. The corresponding identification results are also shown in Table \ref{tab_01} and Fig. \ref{fig_08}.
It should be mentioned that two auxiliary constants $k_a, k_b$ are respectively introduced in place of the terms $\alpha k$ and $(1-\alpha)k$ in CBW model to make the identification more easier. 

It is obviously seen from Fig. \ref{fig_08} that, compared with the CBW model, both the proposed FONBW model and the model in \cite{zhu2016hysteresis} can characterize the asymmetric and rate-dependent hysteresis well. But the predicted result of the FONBW model matches better the experimental response with the RMS error of $0.23\mu$m than that of the model in \cite{zhu2016hysteresis}, which validates the accuracy advantage of the proposed FONBW model.

To further intuitively demonstrate the advantage of fewer parameters in the proposed FONBW model, Table \ref{tab_02} lists a quantified comparison with existing modified BW models \cite{li2016modeling,zhu2016hysteresis,zhu2012non,wang2015modeling,habineza2014multivariable}. Note that although the modified BW models in \cite{wang2015modeling,zhu2012non,habineza2014multivariable} have fewer parameters than the proposed FONBW model, but they only describe the asymmetric hysteresis, which does not meet the requirements of high modeling accuracy in a wide frequency range. 

\begin{table}[!t]
\centering
\caption{Quantified Comparison of Different Modified BW Models}
\label{tab_02}
\begin{tabular}{p{1.8cm}<{\centering} p{3.5cm}<{\centering} p{2.3cm}<{\centering}}
\hline
\hline
Model & Characteristic & Parameters' number  \\ 
\hline
Wang \cite{wang2015modeling} & Asymmetric & 6 \\
Zhu \cite{zhu2012non} & Asymmetric & 7  \\ 
Habineza \cite{habineza2014multivariable} & Asymmetric & 8  \\
Li \cite{li2016modeling} & Asymmetric / Rate-dependent & 22  \\ 
Zhu \cite{zhu2016hysteresis} & Asymmetric / Rate-dependent & 11  \\ 
Proposed & Asymmetric / Rate-dependent & 9  \\
\hline
\hline
\end{tabular}
\end{table}

\subsection{Model Verification} 

\begin{figure}[!t]
	\centering
	\includegraphics[width=3.5in]{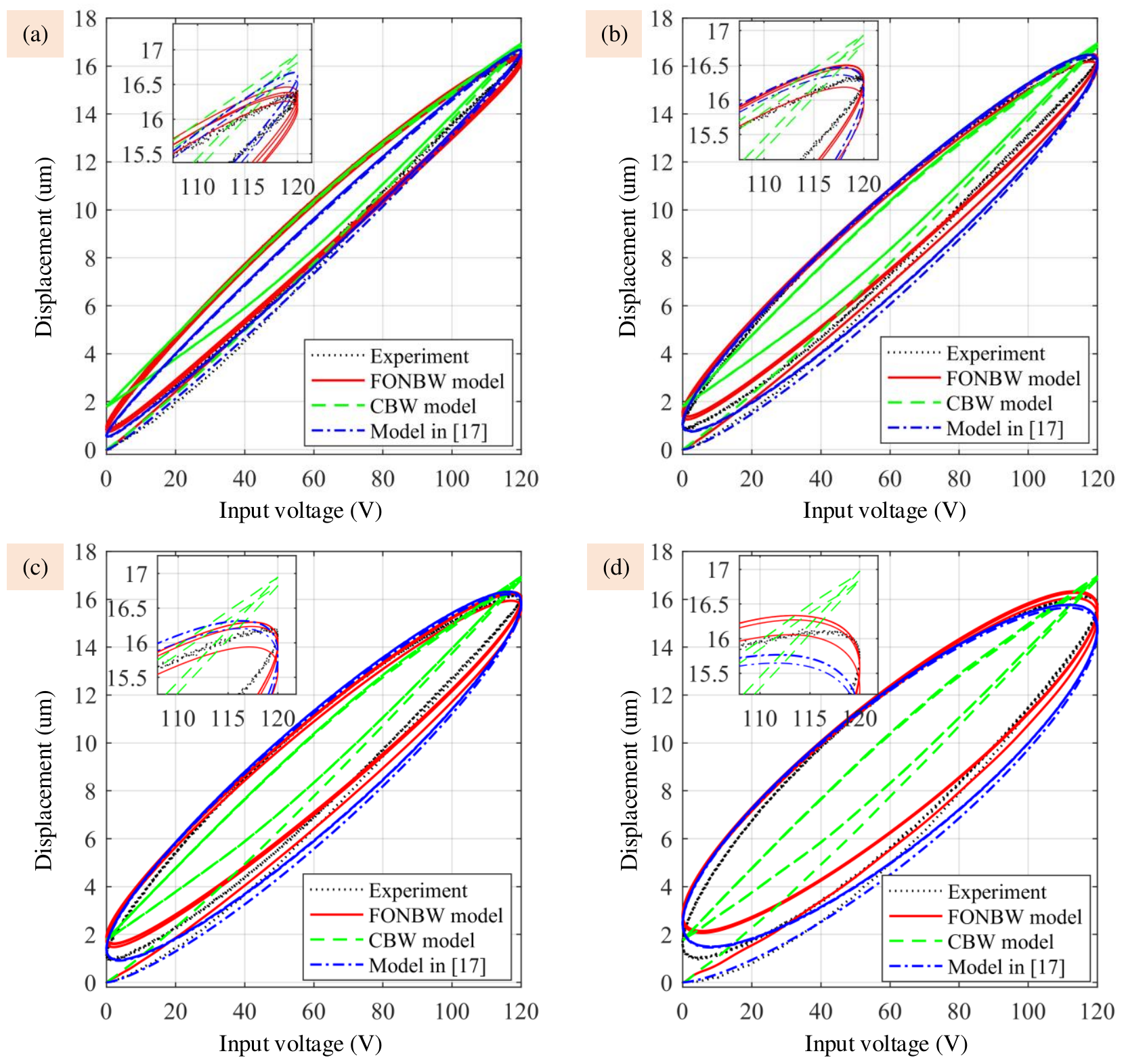}
	\caption{Comparisons of the hysteresis curves of the experimental and the simulation results. (a) $f$=1 Hz. (b) $f$=5 Hz. (c) $f$=10 Hz. (d) $f$=20 Hz.}
	\label{fig_09}
\end{figure} 

\begin{figure}[!t]
	\centering
	\includegraphics[width=3.5in]{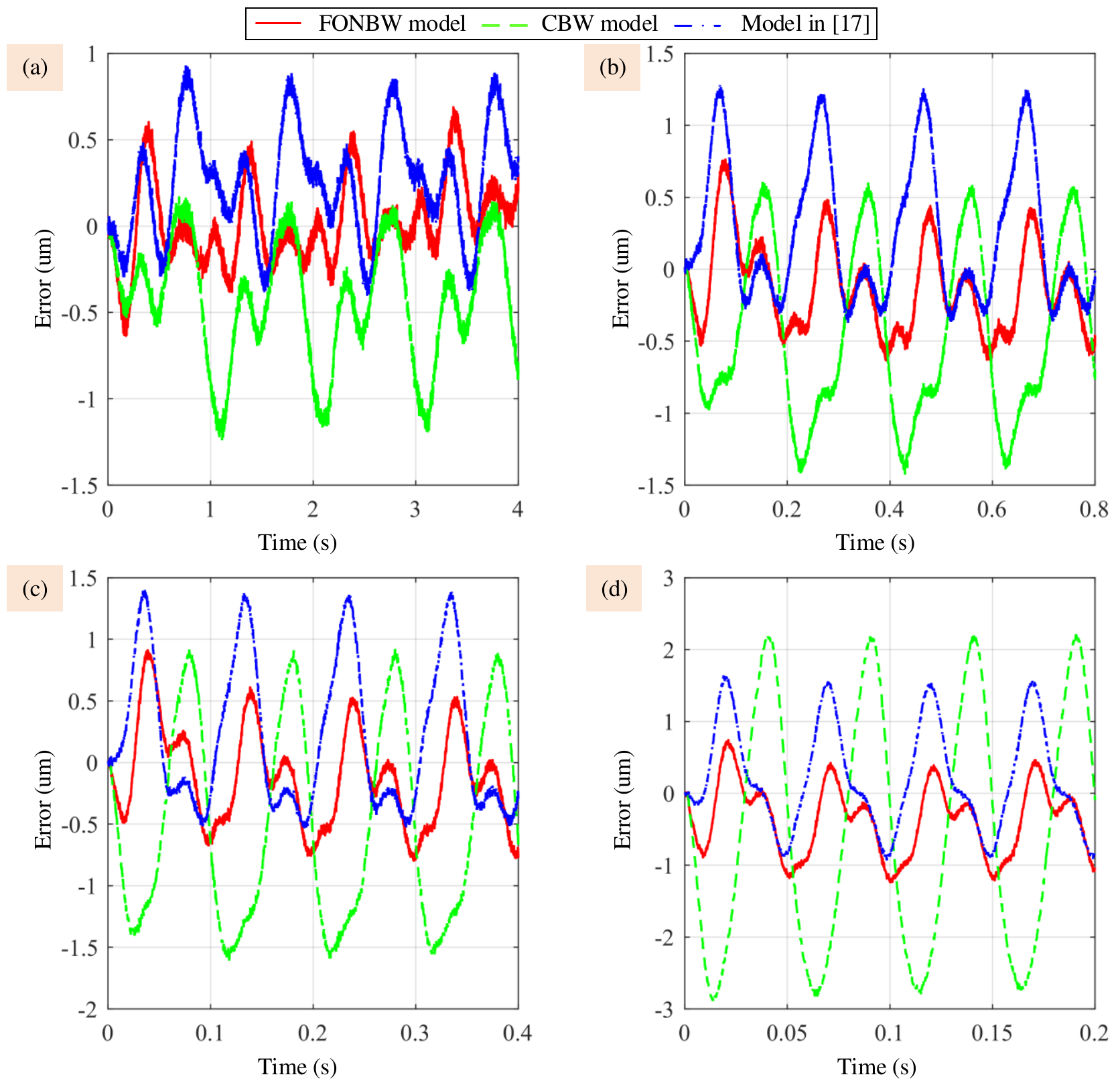}
	\caption{Comparisons of the predicted errors of the proposed model and traditional BW models. (a) $f$=1 Hz. (b) $f$=5 Hz. (c) $f$=10 Hz. (d) $f$=20 Hz.}
	\label{fig_010}
\end{figure} 

To evaluate the effectiveness and feasibility of the identified FONBW model, the sinusoidal input signals $u(t)=\left[ 60-60 \cos(2 \pi f t)\right] $V with different frequency $f$ are applied to the PEA system.

Fig. \ref{fig_09} shows the comparisons of hysteresis curves of the experimental results and the identified models under different input frequencies. The corresponding predicted errors are also demonstrated in Fig. \ref{fig_010}. For the quantified comparison, Table \ref{tab_03} lists the RMS errors of the proposed FONBW model, the CBW model and the model in \cite{zhu2016hysteresis} with respect to the motion range of the PEA system. 

\begin{table}[!t]
	\centering
	\caption{RMS Errors of the Proposed Model and Traditional BW Models With Respect to the Motion Range of the PEA System}
	\label{tab_03}
	\begin{tabular}{p{2cm}<{\centering} p{0.8cm}<{\centering} p{0.8cm}<{\centering} p{0.8cm}<{\centering} p{0.8cm}<{\centering} p{0.8cm}<{\centering}}
		\hline
		\hline
		Frequency & 1 Hz & 5 Hz & 10 Hz & 15 Hz & 20 Hz \\ 
		\hline
		FONBW model & 1.44\% & 2.04\% & 2.54\% & 3.36\% & 3.72\% \\ 
		CBW model & 3.25\% & 4.38\% & 5.61\% & 8.55\% & 10.48\% \\ 
		Model in \cite{zhu2016hysteresis} & 2.44\% & 3.30\% & 3.87\% & 4.13\% & 4.78\% \\ 
		\hline
		\hline
	\end{tabular}
\end{table}

It can be observed that the CBW model cannot exactly describe the complicated hysteresis loops with asymmetric and rate-dependent behaviors. Large errors exist between the model output and experimental results. Specifically, with the increase of the input frequencies, the RMS errors deteriorate sharply from 3.25\% to 10.48\%. In contrast, the FONBW model have a better performance on the asymmetric and rate-dependent hysteresis description. The corresponding RMS errors increase slowly from 1.44\% to 3.72\% in a relative wide frequency band between 1 Hz and 20 Hz. Although the model in \cite{zhu2016hysteresis} also performs the same characteristics as the proposed model, the former has bigger RMS errors than the latter. Therefore, the proposed FONBW model can well solve the asymmetric and rate-dependent behaviors of the piezoelectric hysteresis nonlinearity with fewer parameters and better accuracy, which obviously validates its effectiveness and feasibility.

\subsection{Hysteresis Compensation}

\begin{figure*}[!t]
	\centering
	\includegraphics[width=6.in]{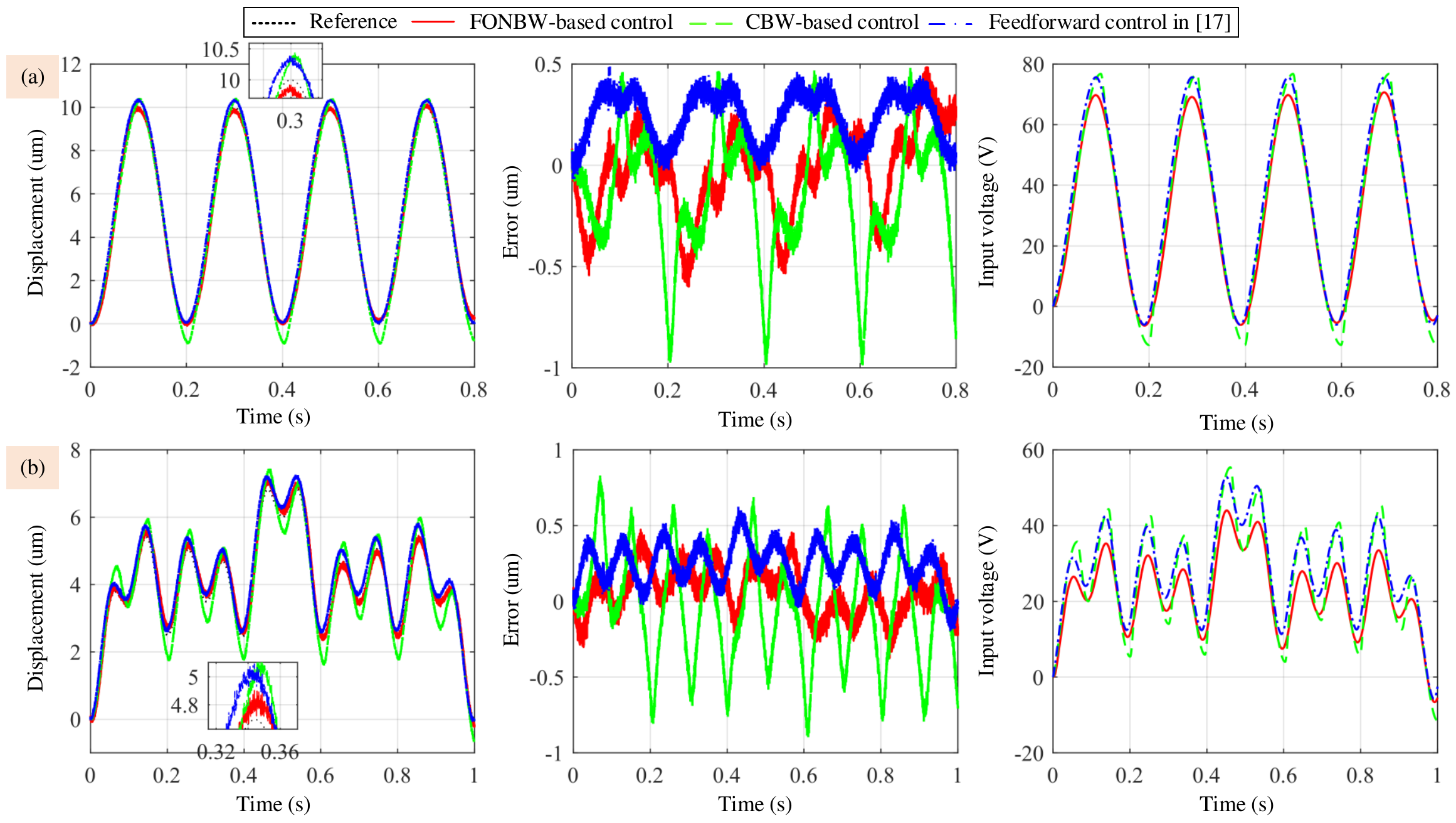}
	\caption{Comparisons of the hysteresis compensation under three feedforward controllers. (a) 5-Hz sinusoidal trajectory. (b) Multi-frequency variable-amplitude trajectory.}
	\label{fig_011}
\end{figure*} 

Furthermore, to validate the hysteresis compensation performance of the proposed inverse-multiplicative-structure-based feedforward control method in Eq. (\ref{eq_016}), the feedforward control experiments are conducted by respectively tracking two kinds of reference signals, i.e., the single-frequency 5-Hz sinusoidal signal $H_d(t)=\left[ 5-5 \cos(10 \pi t)\right] \mu$m and the multi-frequency variable-amplitude signal $H_d(t)=\left[ 4-\cos(2 \pi t) - \cos(6 \pi t)- \cos(10 \pi t)- \cos(20 \pi t)\right] \mu$m. Fig. \ref{fig_011} shows the tracking results, and the corresponding quantitative analysis results are listed in Table \ref{tab_04}, where the CBW-based feedforward control and the feedforward control in \cite{zhu2016hysteresis} are given as follows:
\begin{equation} \label{eq_021}
\begin{cases}
u = \frac{1}{k_a} \left( H_d -k_b D \hat{h}\right)    \\
\dot{\hat{h}} = D^{-1} \left( A \dot{u} - \beta \left| \dot{u} \right| \left| \hat{h} \right|^{n-1} \hat{h} - \gamma \dot{u} \left| \hat{h} \right|^n \right)
\end{cases}
\end{equation}
and
\begin{equation} \label{eq_022}
\begin{cases}
u=\frac{\tau}{k_1 e^{-t/\tau}} \left( m_0 \ddot{H}_d + c_0 \dot{H}_d +k_0 H_d - \hat{h} \right)  \\
\dot{\hat{h}} = A \dot{u} - \beta \left| \dot{u} \right| \left| \hat{h} \right|^{n-1} \hat{h} - \gamma \dot{u} \left| \hat{h} \right|^n + \delta u \mathrm{sgn}(\dot{u}) 
\end{cases}
\end{equation} 
Since the parameters of these two model have been identified in Table \ref{tab_01}, it is not difficult to implement above feedforward control laws.

From the results in Fig. \ref{fig_011} and Table \ref{tab_04}, the CBW-based compensator has the biggest RMS tracking errors among three compensators for both 5-Hz sinusoidal and multiple-frequency variable-amplitude signals, due to its limitation on the hysteresis description. In contrast, the proposed FONBW-based feedforward control performs best with the RMS errors of $0.22 \mu$m and $0.16 \mu$m for 5-Hz sinusoidal and multiple-frequency variable-amplitude signals, respectively. It is seen that the FONBW model is effective for model-based feedforward control.

Besides, the RMS input voltage is also chosen as the performance index. It is found that the proposed FONBW-based control needs smaller control effort than other two control methods, which also demonstrates the advantage of improving control energy of the fractional-order control over the traditional integer-order control.

\begin{table}[!t]
	\centering
	\caption{Hysteresis Compensation Results for Different Desired Trajectories under Three Feedforward Controllers}
	\label{tab_04}
	\begin{tabular}{p{3cm}<{\centering} p{1.5cm}<{\centering} p{1.3cm}<{\centering} p{1.3cm}<{\centering}}
		\hline
		\hline
		Compensator & Trajectory & RMS error & RMS input \\ 
		\hline
		\multirow{2}{*}{FONBW-based control} & 5-Hz &  0.22 $\mu$m & 41.09 V\\ 
		& Multi-freq. &  0.16 $\mu$m & 24.54 V\\
		\hline
		\multirow{2}{*}{CBW-based control}  & 5-Hz &  0.33 $\mu$m & 43.15 V\\ 
		& Multi-freq. &  0.36 $\mu$m & 29.32 V\\ 
		\hline
		\multirow{2}{*}{Feedforward control in \cite{zhu2016hysteresis}}  & 5-Hz  &  0.27 $\mu$m & 44.66 V \\ 
		& Multi-freq.  &  0.26 $\mu$m & 30.45 V\\ 
		\hline
		\hline
	\end{tabular}
\end{table}

\subsection{Discussion}
According to the aforementioned experimental results on the model verification and hysteresis compensation, it has been demonstrated that the proposed FONBW model is effective and superior to the CBW model and traditional modified BW model \cite{zhu2016hysteresis}. The main reason lies in the introduction of the $N$th-order polynomial input function and the fractional-order operators, which makes it possible to describe the asymmetric and rate-dependent hysteresis simultaneously. Moreover, the normalization processing eliminates the redundancy of parameters in CBW model, resulting in fewer parameters than other existing modified BW models with the same characteristics.

However, some limitations still need to be improved in the proposed model. Firstly, the implementation of the fractional-order terms in the model is based on the toolbox FOMCOM \cite{tepljakov2011fomcom} at the Simulink/xPC Target platform. But this toolbox is not available for other general embeded systems. For more practical applications, the numerical calculation needs to be done according to the numerical solutions (\ref{eq_A08})-(\ref{eq_A011}) in Appendix \ref{app_1}.

In addition, it should be noted that in this work, the $N$th-order polynomial input function is adopted as one option to describe the asymmetric behavior. But it cannot be suitable for all cases of piezoelectric hysteresis nonlinearity. Specifically, when the applied input voltage is low (i.e., the hysteresis effect can be ignored.), the behavior of piezoelectric systems comes down to linear. In this case, the $N$th-order nonlinear polynomial input function needs to be replaced with a first-order linear function.

Finally, although the proposed FONBW model can properly describe the piezoelectric hysteresis with asymmetric and rate-dependent behaviors, the modeling errors always exist to some extent as shown in Fig. \ref{fig_010}. Accordingly, the FONBW-based feedforward control will be limited on hysteresis compensation, especially in the presence of parameters variation or external disturbances. Thus, it is certainly better to combine with a robust feedback control method, such as model reference control \cite{kang2020fractional} and sliding model control \cite{li2009adaptive,kang2020model}, to further improve the system's performance.

\section{Conclusion}
\label{sec_5}
In this paper, a new FONBW model is proposed to describe the asymmetric and rate-dependent piezoelectric hysteresis nonlinearity for improving the modeling accuracy. Considering the redundancy of parameters in the CBW model, the normalization processing is firstly conducted such that the model parameters are determined in a unique way. Based on the normalized BW model, an $N$th-order polynomial input function is then utilized to represent the asymmetric hysteresis behavior. Furthermore, two fractional-order operators are introduced into the BW model in place of the integer-order differentials, resulting in the rate-dependent characteristic. Model parameters are identified by the self-adaptive DE algorithm. Comparative experimental results on a PEA system validate the effectiveness and superiority of the developed model on the asymmetric and rate-dependent hysteresis description and the model-based feedforward hysteresis compensation. Thanks to the attractive advantages, the proposed modeling approach provides a wide range of possibilities for the model-based control techniques. Future work will focus on the hysteresis compensation by combining the proposed FONBW-based feedforward control with feedback control approaches to further improve the tracking performance of the piezo-actuated micropositioning systems.

\appendices
\section{Proof of Rate-Dependency}
\label{app_1}
\setcounter{equation}{0}
\numberwithin{equation}{section}

In order to proof the rate-dependency of the proposed FONBW model, it needs to calculate the numerical solution of the fractional-order differential equation (\ref{eq_011}) by making use of the Gr\"{u}nwald-Letnikov definition, which is described as follows:
\begin{definition} \label{def_1} 
	The $\lambda$th-order Gr\"{u}nwald-Letnikov fractional derivative of function $f(t)$ with respect to time $t$ is defined as \cite{monje2010fractional}:
	\begin{equation} \label{eq_A01}
	\mathscr{D}^\lambda f(t)|_{t=p \tau}=\lim\limits_{\tau \rightarrow 0} \frac{1}{\tau^\lambda} \sum_{j=0}^{p} \omega_j^{(\lambda)} f(p \tau - j \tau)
	\end{equation}
	 where $m-1<\lambda<m$, $p, m \in \mathbb{Z}^+$, $\tau$ is the discrete-time step size, and $\omega_j^{(\lambda)}=(-1)^j \binom{\lambda}{j}$, which can be recursively derived by
	 \begin{equation} \label{eq_A02}
	 \omega_j^{(\lambda)}=\left( 1-\frac{\lambda+1}{j}\right) \omega_{j-1}^{(\lambda)}, \; \omega_0^{(\lambda)}=1
	 \end{equation}
\end{definition}

Since the hysteresis effect of piezoelectric material is inherently multi-valued memory-dependent nonlinearity, it is hard to directly obtain the explicit solutions of differential equation (\ref{eq_011}). To facilitate the proof, without loss of generality, one can take $n=1$ for example. Accordingly, Eq. (\ref{eq_011}) is resolved to following four fractional differential equations:
\begin{eqnarray}
\frac{\mathscr{D}^{\lambda_2} \hbar}{\mathscr{D}^{\lambda_1} u} = \rho (1-\hbar)  \qquad \hbar \geq 0, \mathscr{D}^{\lambda_1} u \geq 0 \label{eq_A03} \\
\frac{\mathscr{D}^{\lambda_2} \hbar}{\mathscr{D}^{\lambda_1} u}  = \rho \left( 1+(2 \sigma -1)\hbar \right) \qquad \hbar \geq 0, \mathscr{D}^{\lambda_1} u \leq 0 \label{eq_A04} \\
\frac{\mathscr{D}^{\lambda_2} \hbar}{\mathscr{D}^{\lambda_1} u}  = \rho \left( 1 + \hbar \right) \qquad \hbar \leq 0, \mathscr{D}^{\lambda_1} u \leq 0 \label{eq_A05}  \\
\frac{\mathscr{D}^{\lambda_2} \hbar}{\mathscr{D}^{\lambda_1} u}  = \rho \left( 1- (2 \sigma -1)\hbar \right) \quad \hbar \leq 0, \mathscr{D}^{\lambda_1} u \geq 0 \label{eq_A06} 
\end{eqnarray}

In view of the the Gr\"{u}nwald-Letnikov definition, Eq.(\ref{eq_A03}) can be expanded to be
\begin{equation} \label{eq_A07} 
\tau^{-\lambda_2} \hbar(t) + \mathscr{D}^{\lambda_2} \hbar(t-\tau) = \rho \mathscr{D}^{\lambda_1} u(t) - \rho \hbar(t) \mathscr{D}^{\lambda_1} u(t)
\end{equation}
Combining Eq. (\ref{eq_A07}) with Eq. (\ref{eq_A01}), the numerical solution of Eq.(\ref{eq_A03}) can be expressed by
\begin{eqnarray} \label{eq_A08} 
\hbar(t) = \frac{ \rho \mathscr{D}^{\lambda_1} u(t) -\mathscr{D}^{\lambda_2} \hbar(t-\tau)} {\tau^{-\lambda_2} + \rho \mathscr{D}^{\lambda_1} u(t)} \nonumber \\
= \frac{ \frac{\rho}{\tau^{\lambda_1}} \sum_{j=0}^{t/\tau} \omega_j^{(\lambda_1)} u(t - j \tau) - \frac{1}{\tau^{\lambda_2}} \sum_{i=1}^{t/\tau} \omega_i^{(\lambda_2)} \hbar(t - i \tau) } { \tau^{-\lambda_2} + \frac{\rho}{\tau^{\lambda_1}} \sum_{j=0}^{t/\tau} \omega_j^{(\lambda_1)} u(t - j \tau)} \nonumber \\
, \; \hbar(0)=0 \qquad \qquad \qquad \qquad \hbar \geq 0, \mathscr{D}^{\lambda_1} u \geq 0 \qquad 
\end{eqnarray}

Similarly, Eqs.(\ref{eq_A04})-(\ref{eq_A06}) are also solved and the corresponding numerical solutions are respectively obtained as
\begin{eqnarray} \label{eq_A09} 
\hbar(t) = \nonumber \\
\frac{ \frac{\rho}{\tau^{\lambda_1}} \sum_{j=0}^{t/\tau} \omega_j^{(\lambda_1)} u(t - j \tau) - \frac{1}{\tau^{\lambda_2}} \sum_{i=1}^{t/\tau} \omega_i^{(\lambda_2)} \hbar(t - i \tau) } { \tau^{-\lambda_2} - \frac{\rho (2 \sigma -1)}{\tau^{\lambda_1}} \sum_{j=0}^{t/\tau} \omega_j^{(\lambda_1)} u(t - j \tau)} \nonumber \\
, \; \hbar(0)=0 \qquad \qquad \qquad \qquad \hbar \geq 0, \mathscr{D}^{\lambda_1} u \leq 0 \qquad 
\end{eqnarray}
\begin{eqnarray} \label{eq_A010} 
\hbar(t) = \nonumber \\
\frac{ \frac{\rho}{\tau^{\lambda_1}} \sum_{j=0}^{t/\tau} \omega_j^{(\lambda_1)} u(t - j \tau) - \frac{1}{\tau^{\lambda_2}} \sum_{i=1}^{t/\tau} \omega_i^{(\lambda_2)} \hbar(t - i \tau) } { \tau^{-\lambda_2} - \frac{\rho}{\tau^{\lambda_1}} \sum_{j=0}^{t/\tau} \omega_j^{(\lambda_1)} u(t - j \tau)} \nonumber \\
, \; \hbar(0)=0 \qquad \qquad \qquad \qquad \hbar \leq 0, \mathscr{D}^{\lambda_1} u \leq 0 \qquad 
\end{eqnarray}
and
\begin{eqnarray} \label{eq_A011} 
\hbar(t) = \nonumber \\
\frac{ \frac{\rho}{\tau^{\lambda_1}} \sum_{j=0}^{t/\tau} \omega_j^{(\lambda_1)} u(t - j \tau) - \frac{1}{\tau^{\lambda_2}} \sum_{i=1}^{t/\tau} \omega_i^{(\lambda_2)} \hbar(t - i \tau) } { \tau^{-\lambda_2} + \frac{\rho (2 \sigma -1)}{\tau^{\lambda_1}} \sum_{j=0}^{t/\tau} \omega_j^{(\lambda_1)} u(t - j \tau)} \nonumber \\
, \; \hbar(0)=0 \qquad \qquad \qquad \qquad \hbar \leq 0, \mathscr{D}^{\lambda_1} u \geq 0 \qquad 
\end{eqnarray}

Synthesizing Eqs. (\ref{eq_A08})-(\ref{eq_A011}), it can be seen that computing the solutions of Eq.(\ref{eq_011}) requires not only the current input voltage but also all the history states. This nonlocal memory property of fractional calculus makes it possible to describe the inherent memory effect of hysteresis. Besides, since the step size $\tau$ is usually chosen as the sampling period for practical applications, the time-dependent variation rate of the input signals are also inculded in the solutions, which results in the rate-dependent property of the hysteresis.
Thus, the proof is completed.

%\section*{Acknowledgment}
%The authors would like to thank the Editor and the anonymous reviewers for their valuable comments and kind suggestions to improve the quality of this paper. 

\ifCLASSOPTIONcaptionsoff
  \newpage
\fi



\bibliographystyle{IEEEtran}


\end{document}